\documentclass[letterpaper, 10 pt, conference]{ieeeconf}  

\IEEEoverridecommandlockouts                              

\usepackage{cite}
\usepackage{amsmath,amssymb,amsfonts}
\usepackage{algorithmic}
\usepackage{graphicx}
\usepackage{textcomp}
\usepackage{xcolor}

\usepackage{booktabs}

\usepackage{verbatim}
\usepackage{balance}

\title{\LARGE \bf Automatic Proficiency Assessment in L2
English Learners}

\author{Armita Mohammadi$^1$, Alessandro Lameiras Koerich$^1$, Laureano Moro-Velazquez$^2$ and Patrick Cardinal$^1$%
\thanks{*This work was supported by a MITACS Accelerate grant.}
\thanks{$^{1}$A. Mohammadi, A. Lameiras Koerich and P. Cardinal are with École de Technologie Supérieure, Université du Québec, Montreal, QC, Canada.
        {\tt\scriptsize armita.mohammadi.1@ens.etsmtl.ca, alessandro.koerich@etsmtl.ca, patrick.cardinal@etsmtl.ca}}%
\thanks{$^{2}$L. Moro-Velazquez is with Johns Hopkins University,
        Baltimore, MD, USA.
        {\tt\scriptsize laureano@jhu.edu}}%
}

\begin{document}
\maketitle
\thispagestyle{empty}
\pagestyle{empty}

\begin{abstract}

Second language proficiency (L2) in English is usually perceptually evaluated by English teachers or expert evaluators, with the inherent intra- and inter-rater variability. This paper explores deep learning techniques for comprehensive L2 proficiency assessment, addressing both the speech signal and its correspondent transcription. We analyze spoken proficiency classification prediction using diverse architectures, including 2D CNN, frequency-based CNN, ResNet, and a pretrained wav2vec 2.0 model. Additionally, we examine text-based proficiency assessment by fine-tuning a  BERT language model within resource constraints. Finally, we tackle the complex task of spontaneous dialogue assessment, managing long-form audio and speaker interactions through separate applications of wav2vec 2.0 and BERT models. Results from experiments on EFCamDat and ANGLISH datasets and a private dataset highlight the potential of deep learning, especially the pretrained wav2vec 2.0 model, for robust automated L2 proficiency evaluation.

\end{abstract}

\section{INTRODUCTION}
The accurate assessment of second language (L2) proficiency is crucial in education. Traditionally, this has relied on standardized tests and human evaluation, exemplified by the test of English as a foreign language (TOEFL), the international English language testing system (IELTS), and Cambridge English assessments, which are often costly, time-consuming, and subjective \cite{Chapelle_Douglas_2006}. To overcome these limitations, automated assessment technologies have emerged, aiming for scalable and objective evaluations. Early efforts focused on written language, utilizing rule-based and statistical models \cite{Alderson_2000}. However, spoken proficiency, particularly in spontaneous dialogue, poses a significantly greater challenge due to its inherent variability \cite{Zechner2019}.

Initial automated spoken assessment systems employed feature-based models, extracting handcrafted linguistic and acoustic features via automatic speech recognition (ASR) to quantify proficiency aspects like fluency and pronunciation \cite{Chen_article, Zechner2019}. While systematic, these models struggled with contextual meaning and interactional competence, highlighting the need for methods that capture the dynamic nature of real-world conversations.

Recent advancements in machine learning (ML) and deep learning (DL) have transformed automated language assessment. Neural networks and pretrained models like wav2vec 2.0 \cite{baevski2020wav2vec20frameworkselfsupervised} have shown promising performance in analyzing monologic speech, as seen in tests like TOEFL iBT, Linguaskill Speaking, and the Duolingo English test. However, dialogic speech assessment remains a significant hurdle. Evaluating interactional competence and real-time responsiveness requires models beyond traditional metrics \cite{mcknight_civelekoglu_gales_banno_liusie_knill_2023}.

Current systems primarily focus on structured speech tasks, limiting their applicability to unscripted communication. Deep learning models, while effective in structured speech classification, require further investigation in spontaneous settings. Similarly, adapting text-based models like BERT \cite{devlin2019bertpretrainingdeepbidirectional} for spoken language presents unique challenges due to differences in discourse and tokenization.

This research addresses these limitations through a progressive approach, from structured to dialogic speech, by: (i) evaluating CNNs \cite{oshea2015introductionconvolutionalneuralnetworks}, ResNets \cite{he2015deepresiduallearningimage}, and wav2vec 2.0 for L2 proficiency classification on structured speech from the ANGLISH dataset, (ii) assessing BERT's performance on transcribed speech in a low-resource setting, and (iii) developing methods for evaluating spontaneous dialogue proficiency. In summary, this study provides a comprehensive and robust approach to automated L2 assessment by analyzing fluency and communicative competence in unstructured conversations. 

\textbf{Our main contributions are as follows:}
\begin{itemize}
    \item We evaluate a diverse set of deep learning models for L2 proficiency classification on both structured and spontaneous speech, spanning multiple datasets.
    \item We propose a dialogue-specific preprocessing pipeline to effectively process long-form, multi-speaker audio.
    \item We conduct all experiments under a strict speaker-independent setup to ensure a realistic assessment of model generalizability.
    \item We investigate multi-task learning for the joint prediction of proficiency level and speaker gender, demonstrating improvements over single-task baselines.
    \item We examine how input segmentation strategies influence the performance of text-based classification models.
\end{itemize}

\section{RELATED WORK}
The automated assessment of spoken English proficiency has evolved significantly, driven by the need for scalable and objective evaluation methods. Early approaches, rooted in contrastive linguistics and error analysis \cite{Lado1961, Corder1967}, transitioned toward communicative competence frameworks, emphasizing meaningful communication \cite{Whyte_article, Canale_article, nebhi-szaszak-2023-automatic}. The Common European Framework of Reference for Languages (CEFR) further formalized this shift, integrating error analysis with communicative competence \cite{Joanna_Book, Hawkins_Buttery_2010}.

Initial automated spoken language assessment (ASLA) systems relied on handcrafted features like MFCCs and fluency metrics, trained with SVMs \cite{cortes1995support} and RNNs \cite{zechner-bejar-2006-towards, SpeechRater_article, WANG201847}. These systems demonstrated the feasibility of automating scoring, achieving consistency comparable to human raters. For instance, models proposed by Zechner and Bejar \cite{zechner-bejar-2006-towards} achieved 47.4\% accuracy for five proficiency classes on integrated tasks, surpassing human agreement rates of 43\%. However, they faced limitations due to manual feature engineering and ASR challenges.

Deep learning advancements enabled end-to-end learning from raw data. Models like BD-LSTMs and BD-RNNs improved performance, integrating acoustic and lexical features \cite{Qian2017BidirectionalLF, 10.1145/3459637.3482395}. Recent work highlights the efficacy of self-supervised learning (SSL) models, such as wav2vec 2.0, which outperform traditional methods in spontaneous speech tasks by leveraging raw audio \cite{Bann2022L2PA, 10023019, wrro203353, lo-etal-2024-effective}. Hybrid models, combining SSL, BERT-based text systems, and handcrafted features, have shown the best results, balancing efficiency and nuanced feedback \cite{mcknight_civelekoglu_gales_banno_liusie_knill_2023}.

In text-based assessment, BERT-based frameworks have been explored to enrich contextual embeddings, achieving high accuracy on large datasets \cite{Schmalz2021AutomaticAO, banno-matassoni-2022-cross}. Multimodal approaches, like combining wav2vec 2.0 and Longformer, have further improved dialogic assessment \cite{mcknight_civelekoglu_gales_banno_liusie_knill_2023}. Hierarchical graph-based models and multi-task learning frameworks, which incorporate sentiment analysis and coherence modeling, have also shown promising results \cite{li2024automatedspeakingassessmentconversation, muangkammuen-fukumoto-2020-multi, Yang2022}. Nebhi and Szaszák \cite{nebhi-szaszak-2023-automatic} demonstrated a multimodal multitask transformer model for CEFR-aligned assessment, achieving high accuracy.

While several studies have explored wav2vec 2.0 for L2 assessment, many rely on speaker-dependent setups. Our study specifically emphasizes speaker-independent evaluation, a more realistic and challenging scenario for deploymen

\section{PROPOSED APPROACH}

\subsection{Preprocessing and Segmentation}
Models like 2D CNNs and ResNet require the pre-processing of the speech signal. The preprocessing pipeline consisted of multiple stages, including audio downsampling to 16 kHz, segmentation into 8-sec excerpts for models requiring fixed length inputs, feature extraction using FBanks with 40 Mel-frequency bins, and input normalization using z-score.

\subsection{Dialogue-Specific Preprocessing}
To evaluate the impact of interviewer speech on classification performance, we applied a preprocessing pipeline to separate learner and interviewer segments. Silence was removed using Pydub\cite{robert2018pydub}, with parameters set to a minimum silence length of 500\,ms, a threshold of $-40$\,dB, and a 100\,ms buffer to preserve natural speech flow. Speaker diarization followed a two-step approach: initial segmentation using PyAnnote~\cite{bredin2019pyannoteaudioneuralbuildingblocks} and refinement with SpeechBrain~\cite{speechbrain} and following McKnight et al.~\cite{mcknight_civelekoglu_gales_banno_liusie_knill_2023}, the learner was identified as the speaker with the longest total speaking time. This allowed us to create two dataset versions: one with full dialogues and another containing only learner speech. Since the recordings lacked transcripts, we used Whisper~\cite{radford2022whisper} to generate automatic speech transcriptions and evaluated multiple variants to identify the most accurate for our data. Despite its strong performance, Whisper occasionally produced errors due to overlapping speech and real-world noise conditions. A subset of the data was manually reviewed to verify segmentation accuracy.

\subsection{Models}
\noindent\textbf{2D CNN}: The architecture consists of four convolutional layers with ReLU activation function to introduce non-linearity, thus enhancing the model’s ability to detect robust features. To prevent overfitting and improve model generalization, batch normalization, max pooling, and dropout layers are applied after each convolutional operation.

\noindent\textbf{Frequency-Axis CNN}: Using the same architecture, we apply convolutions specifically along the frequency axis to capture fine-grained spectral features, such as harmonics and fundamental frequency. This enhances the model's ability to detect key audio characteristics while maintaining regularization through batch normalization, max pooling, and dropout.

\noindent\textbf{ResNet}: We employed a 2D CNN architecture with five residual blocks, featuring channel configurations of 128, 128, 256, 256, and 512. Each block uses 2D convolutional layers to process the spectrogram generated by the FBanks. The convolutional layers have a kernel size of 3 and a stride of 2 to downsample the input. The residual connection links the output from the downsampling step to the output of the last convolutional layer within each block. After the convolutional operations, global average pooling is applied before passing the results to a fully connected layer for classification.

\noindent\textbf{wav2vec 2.0}: The models were initialized with a pre-trained model\footnote{https://huggingface.co/patrickvonplaten/wav2vec2-base}. Learners’ audio responses were input to the model, where the transformer encoder generated contextualized
representations. We fine-tuned the wav2vec 2.0 model for our classification tasks. Finally, the encoder’s output was aggregated using a {\it StatisticsPooling} layer with mean pooling, and two linear classifiers produced the final predictions. To enhance generalization and mitigate overfitting, we froze the convolutional feature extractor layer of the wav2vec 2.0 model and fine-tuned its transformer encoder. Separate optimizers were employed for the transformer encoder and the linear classifier, ensuring optimal learning rates for each component. This fine-tuning strategy enables the model to adapt its pretrained representations to the specific requirements of the classification tasks.

\noindent\textbf{BERT}: We explored text-based classification using transcriptions of learner responses. A pre-trained \texttt{bert-base-uncased} model was used to generate contextual embeddings, with its parameters frozen during training to prevent overfitting.

\section{EXPERIMENTAL RESULTS}

\subsection{Datasets}
\noindent\textbf{ANGLISH:} This corpus \cite{Anglish_article} was designed to analyze British English as a second language (L2) among native French speakers. It comprises over 5.5 hours of speech (20,396 seconds) from a variety of structured and spontaneous speech tasks by 67 participants across three proficiency levels: (i) native English speakers (NES), the highest proficiency group, including 13 females and 10 males with an average age of 31 years from various regions in England; (ii) French speakers subdivided into 11 females and 11 males with an average age of 37.5 years without specialized phonetics training (FR1) and university students, 11 females and 11 males, aged between 19 to 22 years with formal phonetics training (FR2). All recordings were conducted in an anechoic chamber to ensure pristine audio quality. We used the reading passages, where participants read four texts, yielding approximately 1.5 hours of speech across 1,260 sentences, and spontaneous monologues, where 63 participants delivered 4-minute talks on pre-suggested topics (e.g., holidays). The audio data was transcribed and segmented at the phoneme level using PRAAT \cite{Boersma2009}, allowing the capture of nuanced linguistic features such as pauses, hesitations, and truncated words. The corpus is publicly available for academic use \cite{bel:hal-00142931}.

\noindent\textbf{EFCamDat:} The EF-Cambridge open language dataset comprises 
1,180,310 essays written by learners and subsequently corrected and evaluated by professional language instructors \cite{Geertzen2014AutomaticLA}. It was created from a global pool of English language learners spanning over 172 nationalities. The dataset represents 16 different proficiency levels mapped to five common European framework of reference for languages (CEFR) levels: A1, A2, B1, B2, and C1, which reflect the progression from beginner (A1) to advanced (C1) proficiency, making the EFCamDat a highly diverse and stratified resource for studying learner language. Each essay is accompanied by rich metadata, including the learner's native language, age, gender, and time spent learning English. 

\noindent\textbf{Private:} The private dataset contains structured interview recordings to assess students' language proficiency. These recordings are dialogue-based, where students respond to interviewer-led questions and proficiency is evaluated across four key dimensions: fluency, comprehension, accuracy, and vocabulary, with an overall score assigned to each recording.  
The dataset is divided into two sections: (i) {\it tests\_linguistiques}, contains learners’ initial proficiency levels, ranging from 1 to 8; 
(ii) {\it tests\_validation} evaluates proficiency after language coursework. 
We used only part of the {\it tests\_linguistiques} section, levels L3 to L5, which data distribution across proficiency levels in show in Table~\ref{LDL}. Such levels provided the largest number of speakers and samples. 

\begin{table}
\centering
\caption{The distribution of data across different levels of the private dataset \label{LDL}} 
		\begin{tabular}{|c|c|c|c|c|}
		\hline
			& \multicolumn{3}{c|}{\bf Duration} & {\bf Number of} \\ \cline{2-4}
            {\bf Level} & {\bf Total} & {\bf Min} & {\bf Max} & {\bf Speakers} \\

	  \hline
			L1 & 3:44:31 & 6:29 & 17:45 & 19 \\
	  \hline
			L2 & 27:03:08 & 4:11 & 25:15 & 135 \\
	  \hline
			L3 & 54:17:21 & 4:06 & 24:25 & 225 \\
	  \hline
			L4 & 50:35:31 & 5:14 & 24:41 & 197 \\
	  \hline
			L5 & 35:50:34 & 6:22 & 27:40 & 126 \\
	  \hline
			L6 & 19:00:50 & 9:08 & 36:14 & 60 \\
	  \hline
			L7 & 9:34:04 & 12:11 & 27:61 & 29 \\
	  \hline
			L8 & 2:06:32 & 19:38 & 33:12 & 5 \\
	  \hline
		\end{tabular}
\end{table}


\subsection{Experimental Protocol}
\noindent\textbf{ANGLISH:} We employed 10-fold cross-validation (CV) using 61 speakers. In each fold, a subset of speakers was used for training, while a distinct subset was held out for validation. Stratification ensured balanced gender and proficiency representation across folds. We also have a test set composed of 6 speakers (one male and one female from each proficiency level), never seen during training or validation, to assess the generalization of the models. 

\noindent\textbf{EFCamDat:} In our experiments, we utilized a carefully curated subset with 2,000 samples for the training set, 200 samples for the validation set, and 200 samples for the test set. To ensure representativeness and fairness, we stratified the subset according to the 16 proficiency levels present in the full dataset. This stratification helps preserve the proportional distribution of samples across levels, enabling a robust analysis of proficiency classification.


\noindent\textbf{Private:} To ensure a fair and robust evaluation, we partitioned the data into training, validation, and test sets. For each level, we randomly selected 12 speakers for the validation and test sets, ensuring that their total audio duration matched 10\% of the total duration in L5, the least represented level. The remaining speakers were used for training. Crucially, we enforced strict speaker separation across partitions, ensuring that no speaker appeared in both training and evaluation sets. This approach guarantees that model performance is assessed on entirely unseen speakers, providing a more reliable measure of generalizability.

\subsection{Implementation Details}

\noindent\textbf{Anglish:} \textit{2D CNN \& Frequency-Axis CNN:} Trained for 200 epochs, batch size 8, learning rate decayed from 0.001 to 0.0001, with early stopping (patience = 10). \textit{ResNet:} 100 epochs, batch size 16, same learning rate schedule. \textit{wav2vec 2.0:} Fine-tuned for 5 epochs, batch size 16, using learning rates of $1\text{e}^{-4}$ (classifier) and $1\text{e}^{-5}$ (encoder), encoder dimension 768. \textit{Multi-task:} A dual-layer DNN (LeakyReLU, batch norm, log-softmax) was used for joint gender (binary) and proficiency (3-class) classification, with loss $\mathcal{L}_{\text{final}} = 3\mathcal{L}_{\text{level}} + \mathcal{L}_{\text{gender}}$. All models used Adam (weight decay = 0.0001).

\noindent\textbf{EFCamDat:} \textit{BERT+SVM:} Fixed BERT embeddings with an RBF-kernel SVM ($C=10$, $\gamma=0.01$), trained via SGD (hinge loss, L2 regularization $= 1\text{e}^{-3}$). \textit{BERT+MLP:} A 3-layer MLP (768, 128, 5) with ReLU and 0.2 dropout, trained for 300 epochs using Adam ($1\text{e}^{-5}$), early stopping (patience = 10). \textit{Fine-tuned BERT+MLP:} Last BERT layers and MLP fine-tuned for 30 epochs (patience = 2). \textit{Token Length:} We tested input lengths from 10 to 90 tokens to evaluate performance–efficiency trade-offs.

\noindent\textbf{Private:} \textit{2D CNN \& ResNet:} Trained for 100 epochs, batch size 16, with learning rate decay (0.001 to 0.0001), and early stopping. \textit{wav2vec 2.0:} Aggregated 2D features (mean pooling), followed by a 768-unit dense layer, 0.2 dropout, softmax output. Trained for 20 epochs using AdamW (batch size 4, learning rate $1\text{e}^{-5}$). \textit{BERT:} Frozen \texttt{bert-base-uncased} with BiLSTM, self-attention, and dense output. Two input formats: (1) 1-min transcript, (2) 7-sentence segments. Trained for 120 epochs (learning rate: 0.001, weight decay: 0.01).

\subsection{Results}
\noindent\textbf{ANGLISH:} Performance was evaluated separately for single-task learning (proficiency or gender classification alone) and multi-task learning (joint proficiency and gender classification). The results for each model under these settings are presented in Table~\ref{tab:modelResults}.

\begin{table}[h!]
    \centering
    \caption{Results of different models on the test set of the Anglish dataset.}\label{tab:modelResults}
    \begin{tabular}{|c|c|c|}
    \hline
        & \textbf{Level ACC} & \textbf{Level ACC} \\
               \textbf{Model} & \textbf{Single-task} &  \textbf{Multi-task}  \\
    \hline
        2D CNN & 20.8\% & 29.2\% \\
    \hline
        Frequency-Axis 2D CNN & 27.4\% & 33.3\% \\
    \hline
        ResNet & 35.9\% & 43.8\% \\
    \hline
         wav2vec 2.0 & 75\% &  80.0\% \\
    \hline
    \end{tabular}
\end{table}

The experimental results reveal significant performance differences across the evaluated models, with pretrained wav2vec 2.0 emerging as the top-performing architecture. Notably, wav2vec 2.0 achieved 75\% accuracy in the single-task proficiency classification task and 80\% in the multi-task setting, demonstrating its superior ability to generalize and capture intricate speech patterns. This performance advantage can be attributed to its pretrained representations, which are adept at encoding complex acoustic features and leveraging large-scale self-supervised learning.




\noindent\textbf{EFCamDat:} Table \ref{EFCAMDAT_table} presents the average lengths of essays across different CEFR proficiency levels, highlighting the variation in text complexity and length as learners progress from A1 to C1. 
\begin{table}[h]
    \centering
        \caption{Level distribution of the EFCamDat dataset with average sequence length}
        \label{EFCAMDAT_table}
    \begin{tabular}{|c|c|c|}
        \hline
        {\bf Level} & {\bf \# of Answers} & {\bf Average Length} \\
        \hline
        A1 & 191,663 & 40 \\ \hline
        A2 & 129,591 & 67 \\ \hline
        B1 & 61,506 & 92 \\ \hline
        B2 & 18,187 & 129 \\ \hline
        C1 & 5,115 & 179 \\ 
        \hline
    \end{tabular}
\end{table}
Table \ref{EFCAMDAT_table} illustrates that lower-level learners (A1, A2) tend to write shorter essays with less variability in length, reflecting their limited vocabulary and writing skills. Conversely, higher-level learners (B2, C1) produce longer and more complex essays with greater variability, indicative of their more advanced language proficiency and diverse use of linguistic structures. This variability underscores the need for adaptive preprocessing strategies, particularly when working with texts of varying lengths.
The performance of different classifiers using a fixed input length of 60 tokens is summarized in Table \ref{tab:bert_model_accuracy}. These results offer insights into the effectiveness of various classification methods applied to BERT embeddings.

\begin{table}[h]
    \centering
\centering
        \caption{Performance of the BERT models with different classifiers and fine-tuned BERT model with max sequence length of 60}
        \label{tab:bert_model_accuracy}
    \begin{tabular}{|c|c|c|c|} 
        \hline
        \textbf{Model Name}    & \textbf{Accuracy} & \textbf{Precision} & \textbf{F1} \\ \hline
        BERT+SVM          & 79.0\%            & 0.798              & 0.791       \\ \hline
        BERT+MLP          & 86.5\%            & 0.88               & 0.89        \\ \hline
        FTBERT+MLP & 87.5\%            & 0.89               & 0.90        \\ \hline
    \end{tabular}
\end{table}

Table \ref{tab:bert_model_accuracy} shows a clear trend of improvement in performance as we move from the simpler BERT+SVM model to the more advanced FTBERT+MLP model. The accuracy, precision, and F1 score all increase with each model, demonstrating the impact of incorporating more sophisticated classification techniques. The comparative analysis of the models reveals distinct performance trends for BERT+SVM, BERT+MLP, and FTBERT+MLP across varying token lengths. The results, summarized in Table \ref{bert_accuracy_table_Various_Token_Lengths}, highlighting the impact of token length on model accuracy and the benefits of fine-tuning.

\begin{table*}
\centering
\caption{Accuracy (\%) of BERT Models at Various Token Lengths}
\label{bert_accuracy_table_Various_Token_Lengths}
    \begin{tabular}{|c|c|c|c|c|c|c|c|c|c|}
    \hline
    & \multicolumn{9}{|c|}{ \textbf{Number of Tokens}} \\ \cline{2-10}
    \textbf{Models} & \textbf{10} & \textbf{20} & \textbf{30} & \textbf{40} & \textbf{50} & \textbf{60} & \textbf{70} & \textbf{80} & \textbf{90} \\
    \hline
    BERT+SVM & 68.5 & 71.5 & 79.5 & 81.0 & 82.5 & 83.5 & 84.0 & 85.5 & 86.0 \\
    \hline
    BERT+MLP & 66.0 & 73.5 & 86.5 & 84.0 & 84.0 & 85.0 & 84.0 & 86.5 & 87.0 \\
    \hline
    FTBERT+MLP & 70.0 & 83.0 & 87.5 & 88.5 & 92.0 & 88.5 & 88.5 & 91.5 & 92.0 \\
    \hline
    \end{tabular}
\end{table*}

\noindent\textbf{Private:} Model performance was evaluated using classification accuracy, macro precision, macro recall, and macro F1-score. Each model was trained on a training set and evaluated on an independent test set. The results, presented in Table \ref{ldl_audio_result}, compare the performance of audio-based models, including CNN, ResNet, and wav2vec 2.0 variants, against text-based models using BERT and highlight the impact of model architecture, input duration, and segmentation strategies on classification performance.

\begin{table}[htbp]
\centering
\caption{Classification accuracy of audio-based and text-based models on the test set of the private dataset}
\label{ldl_audio_result}
\begin{tabular}{|l|c|c|c|c|}
\hline
 &  & \textbf{Macro} & \textbf{Macro} & \textbf{Macro} \\ 
\textbf{Model} & \textbf{Accuracy} & \textbf{Precision} & \textbf{Recall} & \textbf{F1} \\ 

\hline
CNN                                 & 29.2\% & NC      & NC      & NC      \\ \hline

ResNet                              & 31.4\% & NC      & NC      & NC      \\ \hline

wav2vec 2.0 FA 8-sec      & 60.5\% & 0.58    & 0.57    & 0.56    \\ \hline

wav2vec 2.0 FA 30-sec     & 60.0\% & 0.59    & 0.59    & 0.58    \\ \hline

wav2vec 2.0 FA 60-sec     & 62.0\% & 0.61    & 0.63    & 0.61    \\ \hline

wav2vec 2.0 SA 30-sec  & 55.0\% & 0.63    & 0.58    & 0.54    \\ \hline

wav2vec 2.0 SA 60-sec  & 61.0\% & 0.62    & 0.61    & 0.60    \\ 
\hline\hline
BERT FA 1-min & 45\% & 0.49 & 0.45 & 0.43 \\ \hline
BERT FA 7-sent  & 52\% & 0.54 & 0.51 & 0.52 \\ \hline
BERT SA 1-min     & 56\% & 0.54 & 0.52 & 0.52 \\ \hline
BERT SA 7-sent      & 57\% & 0.58 & 0.55 & 0.56 \\
\hline
\multicolumn{5}{l}{FA: Full audio. SA: Student audio. NC: Not computed}
\end{tabular}
\end{table}

The results demonstrate the superiority of wav2vec 2.0 over traditional CNN-based architectures for proficiency classification. Both CNN and ResNet models performed poorly, with accuracies of 29.2\% and 31.4\%, respectively, underscoring their limitations in modeling the temporal complexity of spontaneous speech. In contrast, wav2vec 2.0 models achieved significantly higher performance, with the 8-second full-audio model reaching 60.5\% accuracy and the 60-second variant peaking at 62.0\%. These results suggest that longer audio segments contribute richer contextual cues, although performance gains begin to saturate beyond 30 seconds.

A key insight from the comparison of full-audio (FA) and student-only audio (SA) models is that incorporating the interviewer’s speech improves classification. SA models consistently underperformed, with the 60-second student-only model achieving 61.0\%, compared to 62.0\% for the full-audio counterpart. This highlights the contribution of interactional dynamics—such as turn-taking, interviewer prompts, and response timing—in enhancing audio-based classification.

In contrast, for text-based models, using only the student’s transcript yielded better results. The best-performing text model was BERT SA 7-sentences, with an accuracy of 57\% and a Macro F1 of 0.56, outperforming its full-audio counterpart (BERT FA 7-sentences), which reached only 52\%. This indicates that including the interviewer’s text may introduce noise or irrelevant content, while shorter, student-only segments provide more focused and discriminative information. Notably, reducing input length from one-minute transcripts to 7-sentence excerpts improved results across both FA and SA models.

Overall, audio-based wav2vec 2.0 models outperformed their BERT-based textual counterparts, with the top audio model achieving 62\% accuracy versus 57\% for text. These findings underscore the importance of leveraging speech-specific features for proficiency assessment, while also highlighting modality-specific best practices: include full context in audio but focus on concise, student-only content in text.

\section{Discussion}
Overall, our findings highlight the superiority of wav2vec 2.0 for speech-based proficiency assessment, demonstrating that while text-based models provide useful linguistic insights, they are less effective in isolation. The higher misclassification rates of BERT models, particularly in distinguishing adjacent proficiency levels, suggest that text alone lacks crucial prosodic and fluency-related cues. Future research could explore multimodal fusion strategies, integrating speech and text-based features to capitalize on their complementary strengths. Such an approach has the potential to enhance classification accuracy, particularly for proficiency levels where lexical, syntactic, and prosodic features jointly contribute to speaker differentiation.

Compared to prior work, our results align with studies such as Zhou et al.~\cite{zhou-etal-2019-experiments}, which emphasized the importance of diverse feature representations for proficiency scoring. While their approach relied on manually extracted linguistic features and a Random Forest classifier, our study extends this by leveraging deep learning models that automatically learn high-dimensional representations from speech. Additionally, while Lo et al.~\cite{lo-etal-2024-effective} reported high proficiency classification accuracy using wav2vec 2.0, their results were achieved in a speaker-dependent setting, whereas our study focuses on speaker-independent generalization, a more realistic scenario for proficiency assessment. This distinction is critical as it highlights the limitations of certain state-of-the-art models that may perform well in controlled conditions but struggle with generalization.

Compared to McKnight et al.~\cite{mcknight_civelekoglu_gales_banno_liusie_knill_2023}, which also investigated dialogue-based assessment, our study differs in approach: their work applied regression-based scoring using a wav2vec 2.0 + Longformer model, whereas our study primarily focused on classification-based evaluation. Their results suggest that transformer-based architectures improve scoring accuracy when combined with speech embeddings, yet our findings indicate that wav2vec 2.0 alone struggles when applied to highly spontaneous dialogues, particularly in scenarios with greater fluency variability and disfluencies. This suggests that while transformer-based models are beneficial for structured dialogue-based tasks, more advanced multimodal approaches may be needed to robustly assess conversational speech in real-world settings.

\section{CONCLUSION}

This research explored deep learning methodologies for automated L2 proficiency assessment across speech and text modalities, addressing both structured and spontaneous language contexts. Leveraging novel datasets and state-of-the-art models, we demonstrated the potential of these techniques for robust and scalable evaluation.

For speech-based assessment, pretrained models like wav2vec 2.0 excelled in capturing intricate speech patterns and generalizing to unseen speakers. Multi-task learning enhanced proficiency classification, though it highlighted the need for specialized architectures to balance task complexities. While our speaker-independent approach addressed the limitations of prior work, future research should integrate multimodal features to further improve performance.

In text-based assessment, fine-tuned BERT models significantly improved classification accuracy, confirming the effectiveness of deep contextual representations. We demonstrated strong performance even with limited data and sequence lengths, showcasing feasibility in resource-constrained settings. Future work should explore lightweight transformer variants and optimize sequence length effects for scalable solutions.

Assessing spontaneous dialogues presented unique challenges due to fluency variability and disfluencies. While wav2vec 2.0 and BERT performed well individually, fusion strategies are needed to leverage their complementary strengths. Compared to regression-based approaches, our classification-based evaluation highlighted the need for advanced multimodal models in real-world conversational settings. Future research should investigate multimodal learning and hierarchical modeling techniques for robust dialogue assessment.

Overall, this study underscores the efficacy of deep learning for automated L2 proficiency assessment. Future directions include developing specialized multi-task learning frameworks, optimizing transformer models for resource-limited scenarios, and exploring multimodal approaches for spontaneous dialogue evaluation. These advancements will pave the way for more accurate, efficient, and scalable language assessment systems.


\balance
\bibliographystyle{ieeetr}
\bibliography{root}
\end{document}